\documentclass[lettersize,journal, 10pt]{IEEEtran}
\usepackage{amsmath,amsfonts,amssymb}
\usepackage{algorithmic}
\usepackage{array}
\usepackage[caption=false,font=normalsize,labelfont=sf,textfont=sf]{subfig}
\usepackage{textcomp}
\usepackage{stfloats}
\usepackage{url}
\usepackage{verbatim}
\newtheorem{theorem}{Theorem}
\newtheorem{assumption}{Assumption}
\newtheorem{lemma}{Lemma}
\newtheorem{proof}{Proof}
\newtheorem{corollary}{Corollary}
\usepackage{graphicx}
\usepackage{siunitx}
\usepackage{tikz}
\usepackage{enumitem}
\newcommand*\circled[1]{\tikz[baseline=(char.base)]{
            \node[shape=circle,draw,inner sep=0.8pt] (char) {#1};}}

\def\BibTeX{{\rm B\kern-.05em{\sc i\kern-.025em b}\kern-.08em
    T\kern-.1667em\lower.7ex\hbox{E}\kern-.125emX}}
\usepackage{balance}
\newcommand{\system}[1]{\textit{DeTrigger}}

\begin{document}
\title{DeTrigger: A Gradient-Centric Approach to Backdoor Attack Mitigation in Federated Learning}
\author{Kichang Lee$^1$, Yujin Shin$^1$, Jonghyuk Yun$^2$, Jun Han$^2$, Songkuk Kim$^1$ and JeongGil Ko$^1$ \\ $^1$ School of Integrated Technology, College of Computing, Yonsei University \\ $^2$ School of Electrical Engineering, KAIST} 

\markboth{Journal of \LaTeX\ Class Files,~Vol.~18, No.~9, September~2020}%
{How to Use the IEEEtran \LaTeX \ Templates}

\maketitle

Federated Learning (FL) enables collaborative model training across distributed devices while preserving local data privacy, making it suitable for decentralized systems. However, the decentralized nature of FL also opens vulnerabilities to model poisoning attacks, particularly backdoor attacks, where adversaries implant trigger patterns to manipulate model predictions. In this paper, we propose \system{}, a scalable and efficient backdoor-robust federated learning framework that leverages insights from adversarial attack methodologies. By employing gradient analysis with temperature scaling, \system{} detects and isolates backdoor triggers, allowing for precise model weight pruning of backdoor activations without sacrificing benign model knowledge. We provide a thorough theoretical analysis along with extensive evaluations across four widely used datasets, demonstrating that \system{} achieves up to 251$\times$ faster detection than traditional methods and mitigates backdoor attacks by up to 98.9\%, with minimal impact on global model accuracy. Our findings suggest \system{} to be a robust and scalable solution to protect federated learning environments against sophisticated backdoor threats.

\section{Introduction}
\label{sec:intro}

Federated Learning is a decentralized machine learning approach that trains a global model by aggregating locally trained models from mobile and embedded devices~\cite{mcmahan2017communication, shin2024effective}. This method leverages distributed data and computational resources, reducing the dependency on centralized  processing~\cite{park2023attfl,yao2021fedhm,park2024fedhm}. Federated learning powers various ubiquitous computing applications, such as sensor data analysis~\cite{Park20HeartQuake,shen2024fedconv,ouyang2022clusterfl}, autonomous vehicle~\cite{li2021privacy,pokhrel2020decentralized}, and real-time computer vision~\cite{ahn2023safefac,ouyang2023harmony,deng2023fedinc,yun2024powdew}, by using large, diverse datasets collected from distributed mobile and embedded devices without data sharing. Since such mobile applications often exploit privacy sensitive data~\cite{ra2017smart, lee2025mind}, a key principle of federated learning is preserving local data privacy, as the server aggregates updates without accessing raw data~\cite{mcmahan2017communication,li2021fedmask,t2020personalized}. However, this also means the server cannot verify updates, making federated learning vulnerable to model-poisoning attacks from malicious clients~\cite{bagdasaryan2020backdoor,fang2020local,cao2022mpaf}. Considering a ubiquitous computing environment's distributed and collaborative nature, such malicious participants introduce severe security threats to the federated learning systems~\cite{liu2021distfl,zhang2024sars}.

\begin{figure}[!t]
    \centering
    \includegraphics[width=\linewidth]{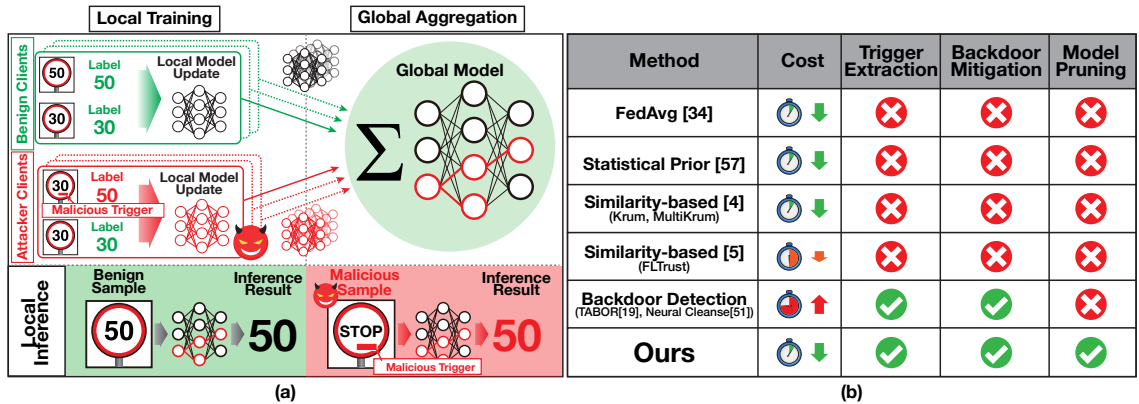}
    \caption{(a) Illustration of backdoor attack in federated learning scenario for local training, server-side global aggregation, and local inference operations. (b) Comparison of our work with previously proposed approaches in addressing backdoor attacks.}
    \label{fig:intro}
\end{figure}

With advancements in federated learning, model poisoning attacks have grown more sophisticated, with the \textbf{\textit{Backdoor Attack}} posing a severe threat~\cite{sun2019can,xie2019dba,li2021invisible,li2021backdoor,yang2025chamaeleon,yang2025catch}. In this attack, as illustrated in Figure~\ref{fig:intro}~(a), an adversary trains a local model to behave benignly on standard inputs but misclassifies inputs with a specific trigger. This compromised model is then submitted to the central server, integrating the malicious knowledge into the global model. Consequently, the global model misclassifies any input containing the trigger, allowing covert manipulation of system outputs. Such attacks are particularly concerning in applications like autonomous driving, where subtle, undetectable modifications to road signs could lead to unsafe decisions and accidents~\cite{li2021backdoor,deng2020analysis,sun2020counteracting}.

Defense mechanisms against backdoor attacks, such as anomaly detection, model filtering, and robust aggregation, aim to neutralize malicious updates by exploiting the statistical prior and behavioral anomalies~\cite{pillutla2022robust,beutel2020flower,sui2026backdoor}. However, as Figure~\ref{fig:intro}~(b) shows, these methods often fail to mitigate backdoor attacks, as backdoor attacks behave normally on legitimate data and activate only on specific triggers~\cite{elhattab2023robust,guo2019tabor}. This makes it difficult to distinguish malicious clients from benign ones without the knowledge of the trigger. Additionally, while some techniques have proven effective in centralized settings, they often entail high computational costs, limiting their scalability in federated learning environments, where servers must assess numerous models each round~\cite{guo2019tabor, wang2019neural}.

We address these limitations by proposing a \textbf{\textit{scalable}} and \textbf{\textit{effective}} strategy for detecting and mitigating backdoor attacks in federated learning, drawing on insights from the relationship between adversarial and backdoor attacks. While adversarial attacks identify perturbation patterns to induce misclassifications at the model, backdoor attacks rely on a hidden trigger pattern embedded in the input to consistently misclassify inputs to a target label. Thus, by employing an adversarial-type approach to design effective noise, we can identify the trigger used in a backdoor attack.

Our proposed mechanism, \system{}, leverages gradient analysis techniques that are widely used for crafting adversarial perturbations~\cite{goodfellow2014explaining,carlini2017towards} for backdoor trigger detection. By examining model gradients, which capture how model weights respond to varying inputs, we can detect subtle deviations indicative of a backdoor trigger. However, this approach presents challenges in distinguishing backdoor attack components from inherent input-specific noise. \system{} addresses this issue by isolating trigger-dependent features within its gradient preprocessing operations. This method is particularly scalable for federated learning, as gradient-based analysis enables efficient abnormal pattern detection without requiring exhaustive inspection of client models on specific triggers. Furthermore, by focusing on gradient behavior, \system{} minimizes the need for extensive verification datasets, which would otherwise be impractical in large-scale distributed systems.

Moreover, \system{} goes beyond simply detecting poisoned models by also isolating the backdoor trigger from the global model. While completely removing compromised models can eliminate backdoor attacks, it also sacrifices the benign knowledge these models contribute. To address this, \system{} leverages the ability to identify the trigger information, enabling the pinpointing and removal of only the backdoor activation weights embedded within the neural network. This approach allows \system{}'s global model to retain the benign knowledge learned from compromised models. Extensive evaluations across four widely used datasets demonstrate that \system{} establishes a backdoor-robust federated learning framework, capable of mitigating up to 98.9\% of backdoor attacks and achieving detection speeds up to approximately 251$\times$ faster than existing methods. This combination of speed and accuracy makes \system{} a scalable and effective solution for securing federated learning.
Specifically, our work makes the following contributions:

\begin{itemize}[leftmargin=*]
    \item We present both theoretical and empirical analyses exploring how leveraging adversarial attack concepts can aid in identifying trigger patterns responsible for backdoor attacks. Our preliminary findings highlight the potential of model gradient analysis for effective backdoor attack detection and mitigation in federated learning.
    
    \item We propose \system{}, a backdoor-robust federated learning framework that combines capabilities for both detecting and mitigating backdoor attacks. By leveraging gradient analysis and insights from adversarial attack methods, \system{} efficiently isolates trigger patterns, allowing targeted removal of malicious activations without sacrificing benign model knowledge. This approach ensures scalable, effective defense against sophisticated backdoor threats in federated learning environments.

    \item We conduct an extensive evaluation of \system{} using four widely used public datasets and various model architectures to demonstrate its scalability and effectiveness in mitigating backdoor attacks. Our results show that \system{} achieves over a 251$\times$ speedup compared to traditional backdoor attack mitigation strategies while preserving the accuracy of the global model and significantly reducing backdoor attack impact.
\end{itemize}


 

\section{Background and Related Work}
\label{sec:relwork}

This section provides background on backdoor attacks in federated learning and an overview of adversarial attacks, emphasizing their similarities, differences, and implications for designing a backdoor-robust framework.

\subsection{Backdoor Attacks}
A backdoor attack aims to force a model $f(\cdot)$ to output an attacker-chosen target label $y_t$ whenever a trigger $T$ is present, while preserving normal predictions on clean inputs. We express a triggered input as
$x_{bd} = (1-M)\odot x + M\odot T$,
where $M$ is a binary mask that controls the trigger location~\cite{jin2022can} and $\odot$ denotes element-wise multiplication. The attack objective is $f(x)=y$ for benign $x$ and $f(x_t)=y_t$ for triggered $x_t$. Backdoors are typically implanted during training by poisoning data or updates, making the malicious behavior persistent and conditionally activated. For example, BadNets inserts simple pixel-pattern triggers into training data to embed backdoors in deep networks~\cite{gu2017badnets}. In traffic sign recognition, an attacker can associate a small physical sticker with a target label (e.g., predicting \emph{stop} regardless of the true sign)~\cite{chen2017targeted}.

\noindent \textbf{Backdoor Attacks in Federated Learning.}
In federated learning, a malicious client can introduce a backdoor by submitting a tampered update for server-side aggregation~\cite{bagdasaryan2020backdoor,gong2022backdoor}. Detection is difficult because the server has no direct access to clients' local data and the backdoor activates only on triggered inputs, so standard validation on clean data may not reveal abnormal behavior.
Prior defenses attempt to mitigate this threat without inspecting client data. Byzantine-robust aggregation reduces the influence of suspicious updates~\cite{fang2020local,cao2020fltrust,zhao2022fedinv}, but it can be ineffective against stealthy backdoors that resemble benign updates. Anomaly detection methods flag abnormal statistics in weights or gradients~\cite{yin2018byzantine,blanchard2017machine}, but they often incur high computational overhead and scale poorly to large models or many clients. Other approaches probe model behavior on synthetic or auxiliary data to detect triggers or backdoor-specific signatures~\cite{elhattab2023robust,wang2019neural,guo2019tabor}. While promising, these methods also face efficiency and scalability limitations, especially in mobile and embedded federated learning settings.

\subsection{Adversarial Attacks}
Adversarial attacks represent a substantial security threat to the robustness of deep neural networks. These attacks are designed to intentionally manipulate the model's output by introducing small, carefully crafted perturbations to the input data. Formally, this can be expressed as $x_{adv} {=} x + \epsilon \cdot n$, $y {=} f(x) {\neq} f(x_{adv})$. In this formulation, \( x_{adv} \) is the adversarial example, created by adding a small perturbation \( \epsilon \cdot n \) to the original input \( x \), where \( n \) is the perturbation direction and \( \epsilon \) controls its magnitude. The goal is to turn the model’s correct prediction (\( y {=} f(x) \)) into an incorrect one (\( y {\neq} f(x_{adv}) \)) while keeping \( \epsilon \) minimal. Research on adversarial attacks focuses on identifying effective perturbations that mislead the model with minimal distortion. For example, Szegedy et al. introduced adversarial examples via L-BFGS optimization~\cite{szegedy2013intriguing}, and Goodfellow et al. later proposed the Fast Gradient Sign Method (FGSM), which efficiently generates adversarial examples by using gradient signs~\cite{goodfellow2014explaining}. These subtle changes, often imperceptible to humans, expose critical weaknesses in neural network robustness~\cite{carlini2017towards}.

\subsection{Relationship Between Backdoor and Adversarial Attacks}
Backdoor and adversarial attacks both seek to undermine a model's reliability by causing incorrect or unintended predictions~\cite{weng2020trade}. In both cases, the attacker perturbs the input distribution seen by the model, either by adding a carefully designed perturbation or by introducing a trigger pattern. Despite this shared objective, the two threats differ fundamentally in when and how the malicious effect is realized.

\textbf{Adversarial attacks} are primarily \emph{inference-time} manipulations. Given a benign input $x$ and a model $f(\cdot)$, the attacker searches for an input-specific perturbation $n$ such that $f(x+\epsilon \cdot n)$ yields an incorrect prediction while keeping $n$ small under a chosen norm constraint ($\epsilon$). This perturbation typically must be recomputed (or at least adapted) for each new input, making the attack inherently \emph{instance-dependent} and requiring continuous interaction with the victim model. \textbf{Backdoor attacks}, in contrast, are \emph{training-time} compromises that implant a persistent association between a trigger pattern and an attacker-chosen behavior. After poisoning the training process, the resulting model behaves normally on benign inputs, yet consistently mispredicts when a predefined trigger is present. Consequently, the attacker does not need to craft a fresh perturbation for each input at inference time; instead, the malicious behavior is activated by a fixed condition (trigger presence). This persistence and conditional activation make backdoor attacks particularly covert, since standard validation on clean data may not reveal any abnormality.

Several defenses for backdoor attacks have been proposed by leveraging insights from adversarial robustness. Jin et al. detect triggers by hierarchically comparing model outputs on benign inputs and adversarially crafted examples~\cite{jin2022can}. Gao et al. show that adversarial training can improve robustness against backdoor behaviors~\cite{gao2023effectiveness}. Wei et al. propose a machine unlearning approach, inspired by adversarial methodologies, to remove backdoor-related vulnerabilities~\cite{wei2023shared}. However, these approaches are not designed for federated learning settings, where scalability across many personalized models, limited compute budgets, and restricted access to centralized clean data can make conventional defenses impractical.
\begin{figure*}[!t]
    \centering
    \includegraphics[width=0.95\linewidth]{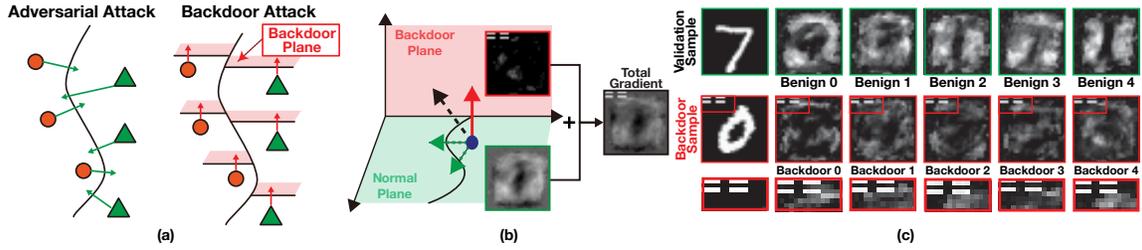}
\caption{(a) Illustration of adversarial and backdoor attacks represented with class decision boundaries. (b) Normal and backdoor-affected gradients for an input sample are presented in the normal data plane and backdoor plane. (c) Samples used in the preliminary study show valid and backdoor samples with detected triggers.}
    \label{fig:core}
\end{figure*}

\section{Core Idea, Feasibility Study, and Analysis}
\label{sec:core}
To counter trigger-based backdoor attacks, identifying the trigger pattern is crucial, as tampered models act benignly without it. This section highlights our defense's core idea to identify the trigger-pattern, inspired by the relationship between the backdoor and adversarial attacks, followed by a feasibility study.

\noindent\textbf{Core Idea.} 
Backdoor and adversarial attacks share overlapping techniques and goals. Adversarial attacks identify subtle input perturbations to alter predictions, while backdoor attacks use a predefined trigger to produce similar incorrect outputs. This raises the question: \textit{“Can adversarial techniques reverse-engineer the backdoor trigger?”} While both share commonalities, they exhibit distinct differences. As shown in Figure~\ref{fig:core} (a), adversarial attacks compute input-specific perturbations, whereas backdoor attacks apply a consistent trigger across inputs. To elaborate, injecting adversarial perturbations into a data sample forces it to shift in an input-specific direction within the representation space, ultimately crossing the decision boundary and altering its prediction. In contrast, a backdoor attack introduces a trigger pattern into each sample, effectively guiding all affected samples toward a consistent direction in the representation space, irrespective of their original features. Notably, a model compromised by a backdoor attack typically learns distinct regions corresponding to the attack target label, forming a backdoor plane in the figure within the representation space. Additionally, adversarial attacks can target any model, while backdoor attacks affect only models trained with trigger samples.


Our proposed framework, \system{}, leverages these distinctions by using adversarial perturbations to identify the backdoor trigger. Figure~\ref{fig:core} (b) presents a high-level illustration of this approach. In computing an adversarial perturbation (i.e., trigger) on a model compromised by a backdoor attack, we hypothesize that the resulting gradients of each client's model capture both input-specific details (dotted green arrows in Figure~\ref{fig:core} (b)) and the backdoor trigger information (solid red arrow in Figure~\ref{fig:core} (b)); sum of these gradients is used for final classification (dashed black arrow in Figure~\ref{fig:core} (b)). This hypothesis is grounded in the properties of gradient-based adversarial perturbations, which are designed to alter the model’s output label. We aim to extract the backdoor trigger pattern by isolating adversarial perturbations through the analysis of input layer gradients.

\noindent\textbf{Feasibility Study.} 
To assess the feasibility of using adversarial techniques for backdoor trigger detection, we conducted a preliminary experiment on the MNIST~\cite{deng2012mnist} dataset using a federated learning setup with 3-layer MLP models. Five benign clients were trained on unaltered data, while five backdoor clients were trained on data with a trigger located in the top-left corner of each image. Each backdoor client applied this trigger to a randomly selected 50\% of their training data. For trigger extraction, we used these 10 client-trained models and computed the input layer gradients using 10 unaltered validation samples excluded from the training set. Input gradients, computed via a single backpropagation pass to the input, capture the sensitivity of the prediction to each pixel. Although they are not meant to reproduce the input itself, they tend to concentrate on localized regions that exert a strong influence on the output. Since backdoor behavior is activated by a fixed, localized trigger, such gradients can provide a convenient signal for revealing the trigger location and pattern.


As illustrated in Figure~\ref{fig:core} (c), the averaged input layer gradients for benign and compromised models reveal a noticeable difference, even though the gradients were generated from an unaltered sample. The backdoor models (i.e., Backdoor 0-4 in the figure) exhibit a distinct trigger pattern (i.e., four white bars) in their input layer gradients. This pattern arises because the gradients consistently emphasize spatial locations or features associated with misclassification across multiple samples. In essence, the backdoor-compromised model has embedded an association between this specific trigger pattern and the target backdoor class in its gradients: offering us hints in effectively extracting the backdoor trigger pattern.

This initial evidence provides the foundation for \system{}, a backdoor-robust federated learning framework. However, despite showing promising initial results, \system{} faces several challenges. First, because the server receives a batch of model updates from its clients, it must efficiently identify the malicious model updates and determine the backdoor’s target attacking label. Moreover, gradient information includes background noise, interpreted as input-specific details, which complicates the accurate extraction of backdoor patterns. The following section will provide a detailed theoretical analysis of the underlying proposed approach. 

{\section{Theoretical Analysis} 
\label{sec:theory_detrigger}
Here, we provide a theoretical justification for \system{}, showing that aggregating target-guided adversarial gradients yields a vector that aligns with the backdoor trigger pattern. Despite input-specific background noise and interference from non-target classes, the empirical mean gradient concentrates around the trigger direction, and mask/energy-based post-processing enables spatial (support) recovery.

\subsection{Definitions and Setup}
Let $x \sim \mathcal{D}_{\text{clean}} \in \mathbb{R}^d$ be a clean input, and assume the data are normalized so that $\mathbb{E}[x]=0$.
Let $y_t$ denote the backdoor target label. The model logit for class $k$ is $z_k(x)$ and the softmax probability is
$p_k(x)=\exp(z_k(x))/\sum_j \exp(z_j(x))$.
Let $T \in \mathbb{R}^d$ be the latent trigger pattern and $M \in \{0,1\}^d$ be a binary mask indicating the trigger support. Define the masked trigger
$T_M \triangleq M \odot T,$ where $\odot$ denotes elementwise multiplication. The masked trigger (backdoored) sample is
\begin{equation}
\begin{aligned}
    x_{bd} &\triangleq (1-M)\odot x + M \odot T, \;\; \\
    &= x + \Delta x,
\end{aligned}
\end{equation}
where $\Delta x \triangleq M \odot (T-x)$. We consider the target-guided adversarial gradient $g(x) \triangleq -\nabla_x \mathcal{L}(x,y_t),$ where $\mathcal{L}$ is the cross-entropy loss.

\subsection{Core Assumptions}

\begin{assumption}[Masked Path-Integrated Dominance and Bounded Curvature]
\label{ass:dominance_curvature}
Consider the linear interpolation path from $x$ to $x_{bd}$:
\begin{equation}
\psi(\alpha) \triangleq x + \alpha \Delta x, \; \alpha \in [0,1].
\end{equation}
Define the sample-dependent backdoor gain in the target logit by the path integral
\begin{equation}
\Delta_{BD}(x) 
\triangleq \int_{0}^{1} \left\langle \nabla_x z_{y_t}(\psi(\alpha)), \Delta x \right\rangle d\alpha.
\end{equation}
We assume the expected gain is positive and substantial: $\mathbb{E}_x[\Delta_{BD}(x)] \gg 0$.
Moreover, the curvature along the trigger direction on this path is bounded by $\kappa>0$:
\begin{equation}
\big|\Delta x^\top H_{z_{y_t}}(\psi(\alpha))\,\Delta x\big|
\;\le\; \kappa \|\Delta x\|^2,
\qquad \forall \alpha \in [0,1],
\end{equation}
where $H_{z_{y_t}}$ denotes the Hessian of $z_{y_t}$ with respect to $x$.
\end{assumption}

\begin{assumption}[Limited Interference from Non-Target Classes]
\label{ass:interference}
For clean samples, the attack target class probability is uniformly small: 
\begin{equation}
    p_{y_t}(x) \le \rho \ll 1.
\end{equation}
Additionally, the absolute interference from non-target classes in the direction $\Delta x$ is bounded in expectation by $\beta$:
\begin{equation}
\mathbb{E}_x \left[
\sum_{j \neq y_t} p_j(x)\,
\left|
\left\langle \nabla_x z_j(x), \Delta x \right\rangle
\right|
\right]
\le \beta.
\end{equation}
\end{assumption}

\begin{assumption}[Signal Consistency and Orthogonal Sub-Gaussian Noise (w.r.t.\ $T_M$)]
\label{ass:signal_noise}
Define the component of $g(x)$ parallel to $T_M$ and its orthogonal residual by
\begin{equation}
c_x \triangleq \frac{\langle g(x), T_M \rangle}{\|T_M\|^2},
\qquad
v_x \triangleq g(x) - c_x T_M.
\end{equation}
By construction, $v_x \perp T_M$. We assume the signal is consistent on average and the residual noise cancels out: $\mathbb{E}[c_x] = \bar{c} > 0,$ $\mathbb{E}[v_x]=0,$ and $v_x$ is a mean-zero sub-Gaussian random vector with parameter $\sigma^2$.
\end{assumption}

\subsection{Lemmas and Main Results}

\begin{lemma}[Lower Bound on the Initial Directional Derivative]
\label{lem:initial_derivative}
Under Assumption~\ref{ass:dominance_curvature}, the expected directional derivative at the clean point ($\alpha=0$) satisfies
\begin{equation}
\begin{aligned}
\mathbb{E}_x\!\left[\left\langle \nabla_x z_{y_t}(x), \Delta x \right\rangle\right]
&\ge
\mathbb{E}_x[\Delta_{BD}(x)]
-
\frac{\kappa}{2}\,\mathbb{E}_x[\|\Delta x\|^2]\\
&\triangleq
\mu_{\text{target}}.
\end{aligned}
\end{equation}
\end{lemma}
In particular, if $\mu_{\text{target}}>0$, then the target-logit gradient exhibits a positive mean projection along $\Delta x$ even at the clean input.

\begin{theorem}[Expected Gradient Alignment (w.r.t.\ $\Delta x$)]
\label{thm:expected_alignment}
Let $g_{\text{adv}} \triangleq \mathbb{E}_x[g(x)]$. Under Assumptions~\ref{ass:dominance_curvature}--\ref{ass:interference},
\begin{equation}
\mathbb{E}_x\!\left[\left\langle g(x), \Delta x \right\rangle\right]
\;\ge\;
(1-\rho)\,\mu_{\text{target}}
\;-\;
\beta.
\end{equation}
Consequently, if $(1-\rho)\mu_{\text{target}} \gg \beta$, the averaged target-guided gradient exhibits a strong positive alignment along the masked-trigger direction of perturbation.
\end{theorem}

\begin{corollary}[From $\Delta x$ to Masked-Trigger Alignment]
\label{cor:delta_to_tm}
Recall $\Delta x = T_M - (M\odot x)$. If the correlation term is negligible,
\begin{equation}
\label{eq:correlation_term}
\left|\mathbb{E}_x\big[\langle g(x), M\odot x\rangle\big]\right| \le \gamma,
\end{equation}
for a small $\gamma\ge 0$ (e.g., due to normalization $\mathbb{E}[x]=0$ and weak dependence between $g(x)$ and $M\odot x$),
then
\begin{equation}
\label{eq:tm_alignment}
\langle g_{\text{adv}}, T_M\rangle
=
\mathbb{E}_x[\langle g(x),T_M\rangle]
\ge
(1-\rho)\mu_{\text{target}}-\beta-\gamma.
\end{equation}
\end{corollary}

\begin{theorem}[High-Probability Convergence of Cosine Similarity]
\label{thm:cosine_convergence}
Let $\hat g = \frac{1}{N}\sum_{i=1}^N g(x_i)$ be the empirical average over i.i.d.\ samples $x_i\sim\mathcal{D}_{\text{clean}}$.
Under Assumption~\ref{ass:signal_noise}, for any $\delta\in(0,1)$, there exist constants $C_1,C_2>0$ such that, for sufficiently large $N$, with probability at least $1-\delta$,
\begin{equation}
\cos(\hat g,T_M)
\;\ge\;
1
-
C_1\cdot
\frac{\sigma^2\,(d+\log(1/\delta))}{N\,\bar c^2\,\|T_M\|^2},
\end{equation}
provided $\hat c \ge \bar c/2$ holds on the same event (e.g., when $c_x$ concentrates around $\bar c$).
\end{theorem}

\subsection{Mask-based Trigger Recovery}
Theorem~\ref{thm:cosine_convergence} guarantees \emph{directional} alignment between $\hat g$ and $T_M$.
For pixel-wise pattern recovery, however, local sign flips and background texture components may obscure direct reconstruction from $\hat g$.
DeTrigger addresses this by using magnitude/energy-based statistics (e.g., $|\hat g|$ or $\hat g^2$) and spatial regularizers (e.g., total variation) to promote a compact, contiguous support, thereby translating directional alignment into accurate mask and pattern recovery.

\section{Threat Model and Assumptions}
\label{sec:threat}
We present the threat model, namely the goal and capability of the attacker, along with the assumptions. 

\noindent\textbf{Threat Model}.
The goal of a backdoor attacker is to manipulate the federated learning process to produce a compromised global model. To achieve this, attackers may alter training data or labels by injecting trigger patterns or tampering with data labels. Also, multiple attackers may collaborate by sharing attack information, including trigger patterns, to increase the chances of compromising the global model. Notably, attackers are limited in interfering with the federated learning process at the local device level. Specifically, they cannot manipulate server-side aggregation, model distribution, client selection, or modify the training processes of other clients.

\noindent\textbf{Assumptions}. We assume that a federated learning system with numerous clients participating without directly sharing their local training data with the central server. Additionally, we assume that malicious clients comprise less than 50\% of the total number of participating clients\cite{fang2020local}. We consider these assumptions reasonable for practical federated learning environments and associated attack scenarios. In addition, we assume that the server has a few clean validation samples that can be used for additional processes~\cite{cao2020fltrust}.

\section{Framework Design}
\label{sec:design}

\begin{figure*}[!t]
    \centering
    \includegraphics[width=0.95\linewidth]{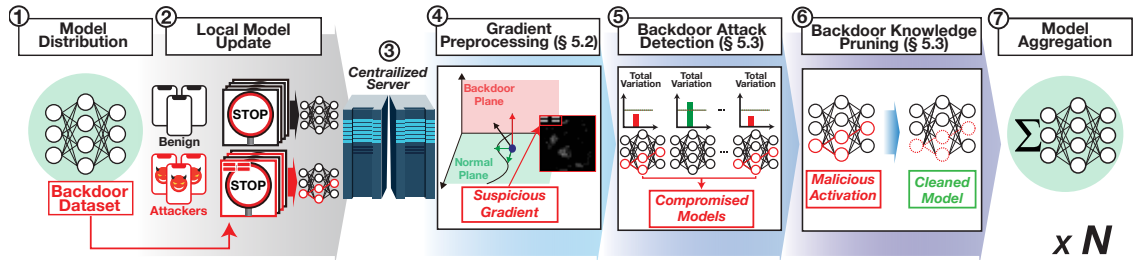}
    \caption{Overall workflow of \system{}. \system{} leverages insights from adversarial attack methodologies to effectively identify the trigger and prune the backdoor knowledge. The server first distributes the global model to selected clients for the local model training. Malicious clients may introduce backdoor triggers into their datasets. To mitigate this, \system{} analyzes gradients to detect potential backdoor triggers and prunes suspicious models before updating the global model.}
    \label{fig:overview}
\end{figure*}

We present the \system{} design, outlining inherent challenges and their corresponding solutions.

\subsection{Overview}
\label{sec:design-overview}

We introduce \system{}, a novel framework for enhancing backdoor robustness in federated learning inspired by principles of adversarial attack-based mitigation. Figure~\ref{fig:overview} presents an overview of \system{} and its operations. \system{} operates as follows: first, the server distributes the latest global model and selects clients for training (\circled{1}). These clients then update their local models using their individual datasets (\circled{2}). If a client is malicious, it may train its model with a backdoor dataset with trigger patterns embedded. After local training, clients send updated model weights back to the server (\circled{3}). At this point, the server computes input layer gradients using a small validation dataset (10-1000 samples) across all labels to evaluate each client’s updated model and \system{} preprocesses model gradients to extract potential backdoor triggers (\circled{4}) (c.f., Sec.\ref{sec:design-gp}). Operating under the insight that the trigger patterns can be reviled when analyzing the input layer gradients (c.f., Sec.~\ref{sec:design-bad}), the server identifies suspicious model updates that may contain backdoor knowledge. When detected, \system{} tests suspicious models with data containing the inferred trigger. If predictions from the model change due to the trigger, \system{} flags the model as compromised (\circled{5}) (c.f., Sec.\ref{sec:design-bad}).

To neutralize backdoor knowledge and create a compromised-but-clean model, \system{} prunes malicious activations in the compromised models by closely observing how their neural network gradients respond to backdoor-embedded samples (\circled{6}) (c.f., Sec.~\ref{sec:design-bkp}). Finally, \system{} aggregates both benign and ``cleaned'' malicious model updates, producing a refined global model (\circled{7}). The following sections detail each core operation of \system{} in depth.

\subsection{Gradient Preprocessing}
\label{sec:design-gp}
\begin{figure}[t!]
    \centering
    \includegraphics[width=0.95\linewidth]{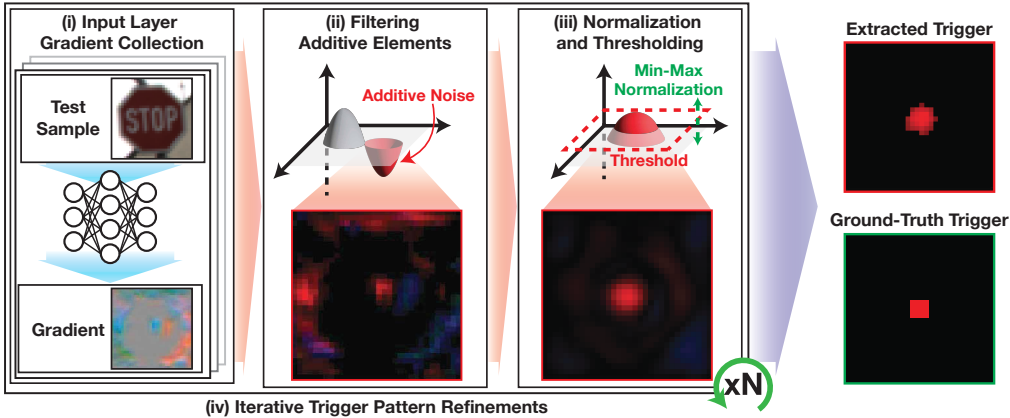}
    \vspace{-2ex}
    \caption{Illustration of gradient preprocessing and trigger extraction operations in \system{}.}
    \label{fig:gp}
   \vspace{-2ex}
\end{figure}

\noindent \textbf{Trigger Extraction.} Extracting backdoor trigger patterns based on gradient information presents challenges as gradients (at the input layer) inherently contain biases specific to the input data, limiting their representation of general patterns. Consequently, raw gradients include both trigger-related information and input-specific noise. The objective of gradient preprocessing is to address these challenges by isolating (and identifying) the backdoor trigger information from irrelevant details embedded in the raw gradients calculated using validation data samples. Gradient preprocessing in \system{} consists of four key steps: (i) input layer gradient collection (ii) filtering additive elements, (iii) normalization and thresholding, and (iv) iterative trigger pattern refinement. We illustrate these operations in Figure~\ref{fig:gp}.

To identify and extract an accurate trigger pattern, \system{} starts by utilizing an uncontaminated validation dataset (covering all labels) at the server to \textit{extract the input layer gradients}. Here, for each model collected from the clients, each sample in the validation dataset is passed through the model, followed by a single round of backpropagation to capture the input layer gradients. As a result, an input layer gradient is recorded for each sample-model pair across all labels in the system.

Using these input layer gradients, \system{} \textit{identifies and filters additive elements} within each model’s gradient that shift predictions from the original to the target attack label. As also demonstrated in our preliminary studies (Figure~\ref{fig:gp}~(c)), this process extracts information within the gradients specifically linked to the trigger pattern. Since the fundamental concept of the backdoor attack involves "injecting" a malicious trigger pattern to manipulate the model into producing an incorrect prediction, \system{} selectively utilizes only the elements that exhibit positive perturbations in shifting the original prediction toward the target label. As shown in Figure~\ref{fig:gp} (ii), the filtered gradient not only includes the backdoor trigger-related feature (red pixels in the center) but also input-specific features. To minimize sample-specific noise variations, \system{} averages the gradient elements across multiple samples used for extracting the input layer gradient. The averaged gradients are then \textit{normalized} through min-max scaling, and a mask is generated to emphasize spatial locations where gradient amplitudes exceed a set threshold. As the sample depicted in Figure~\ref{fig:gp} (iii), this averaging and thresholding process allows \system{} to focus on regions with stronger trigger-related signals, effectively filtering out irrelevant information. In this paper, we set the threshold to 0.5.

Finally, \system{} refines the trigger pattern by modifying test samples, and \textit{iteratively refines the trigger quality}. Specifically, it replaces pixels within masked locations with processed gradients and recalculates adversarial perturbations anew. While adversarial attacks \textit{add} perturbations to original input data, backdoor attacks \textit{replace} specific pixels with the trigger pattern. This iterative process reduces input-specific biases, enhancing the clarity of the trigger pattern by adhering to the foundational concept of backdoor attacks.
Through this process, \system{} generates potential trigger patterns for each model update and label, resulting in \( K \times C \) potential trigger patterns, where \( K \) and \( C \) represent the number of labels and updated clients, respectively.

\begin{figure}[!t]
    \centering
    \includegraphics[width=.85\linewidth]{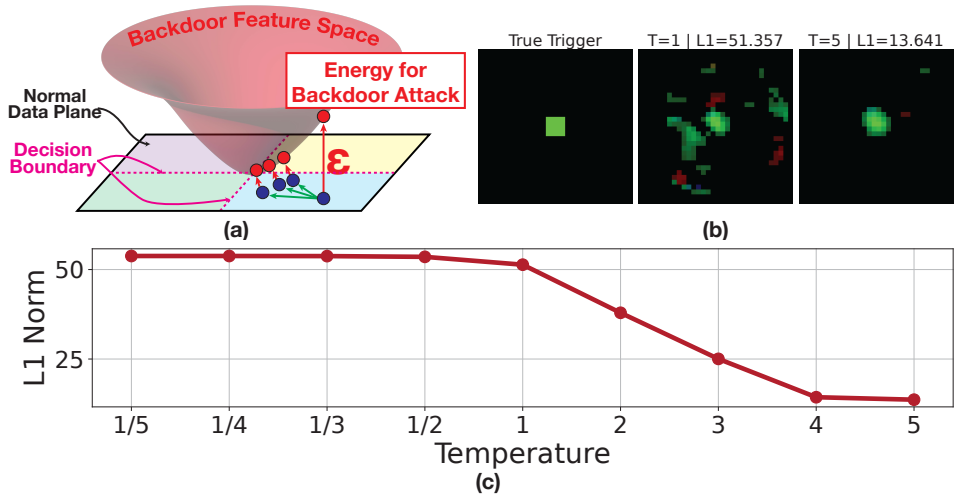}
\caption{(a) Conceptual illustration of the impact of temperature scaling on normal data feature space and backdoor feature space. (b) Sample of ground truth and inferred triggers with different temperatures. (c) L1-norm between ground truth and inferred triggers with varying temperatures.}
    \label{fig:temptrick}
\end{figure}

\noindent \textbf{Temperature Scaling.}
Additionally, as mentioned above, gradients often include input-specific information that is not related to the backdoor trigger, which complicates their extracting process. Here, to better obtain the trigger-related information from the gradient, we introduce {\it temperature scaling}-based trigger-related information amplification during the gradient calculation operations.
Note that the temperature parameter \( T \) modulates the smoothness of the probability distribution output by the softmax function~\cite{hinton2015distilling, lee2024flex}. Higher values of \( T > 1 \) smooth the probability distribution, while values between 0 and 1 sharpen it. This smoothing can be understood as effectively moving a data sample closer to the decision boundary, we note that this can potentially enhance the focus on backdoor-relevant features. 

To elaborate, Figure~\ref{fig:temptrick} (a) visualizes a backdoor model’s decision boundary, illustrating a backdoor feature space lying near the decision boundary intersection within the normal data plane (red cone shape). Given the complexity and high dimensionality of neural network feature spaces, we hypothesize that benign and backdoor models share a ``normal data plane'' where standard samples are positioned, while backdoor samples exist near, but just beyond, this plane. We hypothesize a cone shape for the backdoor feature space as the probability of a sample falling into this space increases as samples are closer to the intersection of decision boundaries in the normal data plane. 


We note that this presumption aligns with the previous theoretical analysis. To elaborate, temperature scaling ($\tau>1$) smooths the softmax distribution, reducing extreme probability imbalances across non-target classes.
In our analysis, this effectively suppresses the interference term (captured by $\beta$) and stabilizes low target confidence (small $\rho$),
thereby increasing the alignment margin $(1-\rho)\mu_{\text{target}}-\beta$ and enabling clearer recovery with fewer samples $N$.

Figures~\ref{fig:temptrick} (b) and (c) illustrate inferred trigger patterns and the impact of temperature scaling on the L1-norm, respectively. Specifically, the results show the L1-norm of gradients decreases with moderate temperature scaling (\( T > 1 \)), improving the clarity of inferred trigger patterns. We observe that increasing the temperature shifts the sample’s position within the normal data plane closer to the decision boundary while simultaneously reducing the influence of irrelevant features (i.e., input-specific noise), as illustrated in Figure~\ref{fig:temptrick} (b). This phenomenon implies that a higher temperature reduces the required perturbation magnitude (\(\epsilon\) in Figure~\ref{fig:temptrick} (a)) to alter the decision toward the backdoor attack target. Consequently, this supports our hypothesis that the backdoor feature space resides near the decision boundary within the normal data plane. This insight aligns with prior findings by Su et al., who observed that the decision boundary of the backdoor sample is tangent to other labels~\cite{su2024model}. This suggests gradients obtained with smoothed probability distributions (i.e., higher temperatures) can more effectively capture information related to the backdoor trigger. Note that we set the temperature to five in this work.

\subsection{Backdoor Attack Detection Module}

\label{sec:design-bad}

Given the trigger pattern extracted above, the next step is to identify backdoor-affected models among those transmitted from clients. Furthermore, we should determine the target label/class used for the attack. The operation consists of two steps, namely, Total variation-based contaminated model detection and Transferability-based verification. 

\noindent \textbf{Total variation-based contaminated model detection.} In this step, \system{} leverages the prior knowledge that backdoor trigger patterns are typically more spatially concentrated than standard adversarial perturbations~\cite{wang2019neural}. To detect the backdoor-affected models and pinpoint the target label, \system{} evaluates the \textit{total variation} for potential trigger patterns. The total variation ($TV$) of an input layer gradient map (i.e., the map of gradients as in Figure~\ref{fig:temptrick}~(c)) is defined as the sum of the absolute differences between neighboring elements along both the horizontal (\(x_{i, \cdot}\)) and vertical (\(x_{\cdot, j}\)) axes in the data (\(x\)). For example, given a gradient map for an input sample, we compare each element by shifting the map both horizontally and vertically by one step. The $TV$ helps identify concentrated spatial patterns indicative of backdoor triggers. Specifically, \system{} computes the total variation 
of the preprocessed gradients for all potential attack target labels across each model as follows: $TV(x) = \sum_{i,j} (\left| x_{i+1, j} - x_{i, j} \right| + \left| x_{i, j+1} - x_{i, j} \right|)$. 
\begin{figure}[t]
    \centering
    \includegraphics[width=.85\linewidth]{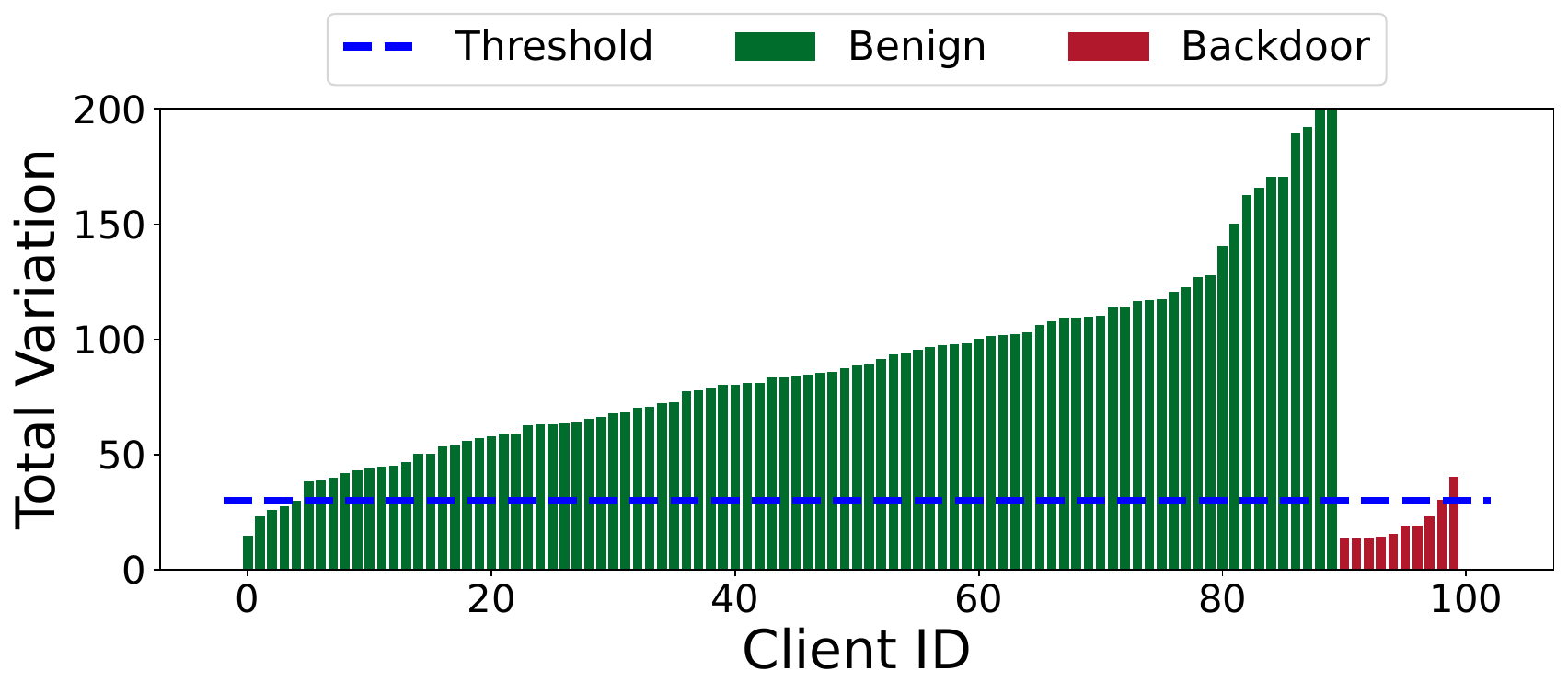}
    \caption{Minimum total variation $TV$ of the processed input layer gradients across different clients. Note the lower $TV$ trend for backdoor-affected gradients.}
    \label{fig:tv-client}
\end{figure}

Using this, if $TV$ falls below a predefined threshold, the evaluated model is flagged as \textit{suspicious}, suggesting it could be influenced by a backdoor, and the associated label is designated as a possible target label. We make this design choice given that backdoor trigger patterns will be spatially dense and having such elements will decrease the $TV$. Note that, the threshold for determining this is adaptive to input data dimensions and resolution, enabling the server to set appropriate thresholds by computing gradients based on centrally available information.

\noindent \textbf{Transferability-based Verification.} Nevertheless, \system{} must address the possibility of \textit{false positives} (i.e., benign clients identified as attackers) and \textit{false negatives} (i.e., attackers classified as benign) in the detection process. Figure~\ref{fig:tv-client} presents the minimum $TV$ values across clients, showing that while backdoor updates (c.f., red bars) typically yield lower $TV$ values than most benign updates, certain benign updates occasionally fall below the threshold, and some backdoor updates exceed the threshold due to data heterogeneity within the federated learning scenario.

To address false detections, \system{} verifies the \textit{transferability} of inferred trigger patterns. First, it examines suspicious models, which may include benign ones, using the potential backdoor triggers extracted from $TV$ thresholding. By injecting these triggers into the validation data and observing if predictions shift towards the target label (instead of the ground truth), \system{} identifies specific models as malicious. If no classification errors are seen, the model is removed from suspicion. This process allows \system{} to effectively isolate trigger patterns that activate backdoors. Subsequently, \system{} tests any remaining models with these verified trigger patterns, flagging any additional models that respond to the trigger as adversaries. 

\subsection{Backdoor Knowledge Pruning Module}
\label{sec:design-bkp}


To motivate our model pruning approach, we ask the following question: \textit{``Can we simply discard malicious updates if we identify them as suspicious?''} To answer this, we set up a motivating experiment using the MNIST dataset with 10 clients: five benign, and five backdoor attackers. Figures~\ref{fig:prune-motiv} (a) plots the data distribution for each client, where the circle size represents the relative sample count per label. We also present model accuracy results for different data distributions in Figure~\ref{fig:prune-motiv} (b). Here, we aim to mimic a scenario where benign clients lack samples from labels 7-9, while attackers possess this data; representing a realistic setup in mobile and embedded applications where data distribution is often highly heterogeneous.

In our experiment, we compare two federated training approaches: FedAvg~\cite{mcmahan2017communication}, which simply averages all client updates, and an Oracle, a hypothetical server with perfect attacker knowledge, which completely excludes malicious models as a whole from aggregation. As shown in Figure~\ref{fig:prune-motiv} (b), the Oracle effectively mitigates the backdoor attack, with only a 2.02\% backdoor attack success rate, as malicious model updates were fully discarded. In contrast, FedAvg shows a higher backdoor attack success rate of 50.25\% since it does not mitigate the attack in any way. Nevertheless, the Oracle exhibits a significant accuracy drop on data for labels 7-9 (37.53\%) due to the loss of benign knowledge embedded within malicious updates, while FedAvg maintains 90.07\% accuracy on these labels. These results suggest that entirely discarding malicious models can result in a severe performance loss for rare classes in heterogeneous data distributions and highlight the importance of leveraging benign knowledge within malicious models while effectively mitigating backdoor attacks, especially in mobile environments where data heterogeneity is common.

\begin{figure}[!t]
    \centering
    \includegraphics[width=1\linewidth]{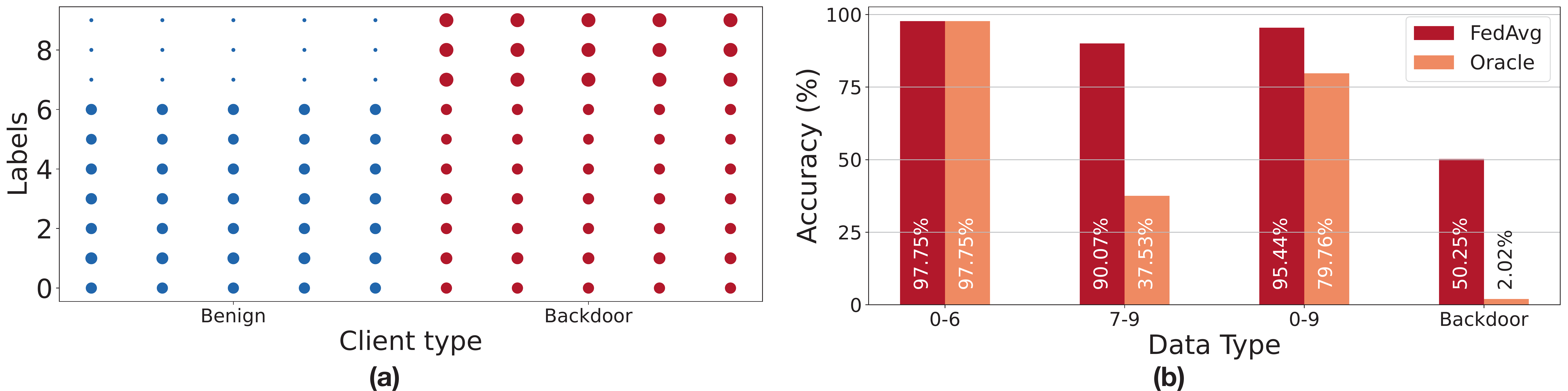}
\caption{(a) Data distribution used in the preliminary motivational study. (b) Model accuracy for different data labels with backdoor attack success rates.}
    \label{fig:prune-motiv}
\end{figure}

We tackle this issue by proposing a \textit{backdoor knowledge pruning module} that only eliminates model parameters associated with backdoor triggers: preventing global model contamination while preserving beneficial knowledge in the model aggregation process. For this, \system{} exploits the backdoor-affected models, the extracted triggers, and target labels as identified in the backdoor detection phase. 

Based on the observation from previous work that backdoor triggers activate distinct weights compared to benign samples~\cite{gu2017badnets}, a \textit{weight pruning} technique that removes these specific weights can be a suitable approach for mitigating backdoor attacks. 
Specifically, the extracted trigger pattern is fed into the malicious model to identify weights that contribute to activating the backdoor. We rank weights based on their gradient norm, pinpointing the contributing weights based on the gradients as contributors to the backdoor activation. These high-gradient weights are then replaced with zeros to neutralize the backdoor effect. Once pruning is completed, \system{} aggregates the pruned models through weight averaging, while the model aggregation approach can be system-specific. This pruning approach allows \system{} to effectively reduce backdoor contamination without discarding the benign knowledge embedded within the malicious client models, enabling more resilient and accurate global model performance across heterogeneous data distributions.



\section{Evaluation}
\label{sec:eval}

We now evaluate \system{} using extensive experiments with four datasets and various comparison baselines.

\subsection{Experiment Setup}

The details on the datasets and models that we use in our evaluations are presented below:

\vspace{0.1in}
\noindent\textbf{Dataset and Model.}
In this work, we evaluate \system{} using four distinct datasets and two model architectures suitable for mobile/embedded federated learning: a 2-layered CNN and ResNet18. We detail the datasets and models below.

\noindent $\bullet$ \textbf{CIFAR-10/CIFAR-100~\cite{krizhevsky2009learning}} are widely used image benchmark datasets, each with 60,000 images at 32$\times$32 resolution, covering 10 and 100 classes, respectively. We employ a 2-layered CNN model as default~\cite{mcmahan2017communication}. Given CIFAR-100’s broader label set, it serves as an ideal dataset for testing \system{}'s adaptability to an increased number of classes.

\noindent $\bullet$ \textbf{GTSRB~\cite{Houben-IJCNN-2013}} is a benchmark dataset comprising 43 types of real-world traffic signs. Given the practicality and vulnerability of traffic sign recognition to backdoor attacks, this dataset enables us to assess \system{} under realistic backdoor scenarios with the 2-layered CNN model.

\noindent $\bullet$ \textbf{STL-10~\cite{coates2011analysis}} contains a total of 13K natural images across 10 classes, with a resolution of 96$\times$96 pixels. The higher resolution of STL-10, compared to the other datasets, makes it suitable for evaluating \system{}'s scalability concerning image resolution. For this dataset, we utilize ResNet18~\cite{he2016deep} to handle the increased complexity.

\vspace{0.1in}
\noindent\textbf{Baselines.} To represent the worst and ideal cases, we use FedAvg~\cite{mcmahan2017communication} and an Oracle configuration. Specifically, FedAvg represents a naive approach, where all model updates are aggregated without any defense against backdoor attacks. Contrarily, the Oracle assumes an ideal setting with complete knowledge of attacker clients, enabling selective exclusion of their updates to prevent contamination of the global model.

To leverage secure aggregation using statistical priors, we also compare with the Median and Trimmed Mean aggregation approaches~\cite{yin2018byzantine}. Median aggregation computes the median rather than the mean, and Trimmed Mean discards outlier parameters with extreme values before averaging updates. These Byzantine-robust methods are effective against straightforward model poisoning, given that malicious updates often deviate significantly from benign ones.

We also compare \system{} with three other federated frameworks that assess update similarity. Krum~\cite{blanchard2017machine} identifies reliable updates by calculating pairwise Euclidean distances and selecting the update with the minimum total distance. MultiKrum~\cite{blanchard2017machine} enhances this by discarding a portion of the most distant updates before averaging, improving resilience against outliers. Lastly, FLTrust~\cite{cao2020fltrust} trains a verified model on the central server and computes trust values based on the cosine similarity between client model updates and the verified model. These trust values adjust aggregation ratios to suppress potential malicious updates effectively. 

These baselines aim to suppress the effect of suspicious clients rather than accurately identifying the backdoor attacker and trigger information. To highlight the practicality of \system{}, we compare it with advanced backdoor defense methods such as Neural Cleanse~\cite{wang2019neural} and TABOR~\cite{guo2019tabor}, which are specialized in backdoor detection and trigger identification. While Neural Cleanse and TABOR excel in identifying triggers through optimization-based approaches, their computational inefficiency makes them unsuitable for federated learning scenarios. This comparison underscores the advantage of our framework in achieving robust backdoor defense without compromising efficiency.

\vspace{0.1in}
\noindent\textbf{Attack Method.} In this work, we address patch-trigger-based backdoor attacks, where attackers train models using backdoor samples containing a designated patch overlaid on clean images. Unless otherwise specified, we use a red square patch sized at \(\frac{1}{8}\) of the image resolution. For example, we apply a 4\(\times\)4 patch for datasets like CIFAR-10, CIFAR-100, and GTSRB, which have 32\(\times\)32 resolution, and a 16\(\times\)16 patch for the STL-10 dataset with a 96\(\times\)96 resolution. Later in our evaluations, we present the performance of \system{} using different trigger patterns as well.

\vspace{0.1in}
\noindent\textbf{Miscellaneous Configurations.} Regarding the federated learning process, unless mentioned otherwise, we split the datasets across 100 clients, which include 25 backdoor attackers with non-independent and identical label distribution following a Dirichlet distribution with the importance parameter $\alpha=$0.5~\cite{hsu2019measuring}. The backdoor attackers generate a backdoor training set by injecting trigger patterns into 25\% of the training data. For every federated training round, 10 local models were selected and trained for 5 epochs with the Adam optimizer, with a learning rate of 5e-3 and a batch size of 64. Throughout the evaluation, we used a server with an Nvidia RTX 3090 GPU, an Intel Xeon Silver 4210 2.20GHz CPU, and 128GB RAM.

\subsection{Overall Performance}
\begin{figure}[!t]
    \centering
    \includegraphics[width=0.95\linewidth]{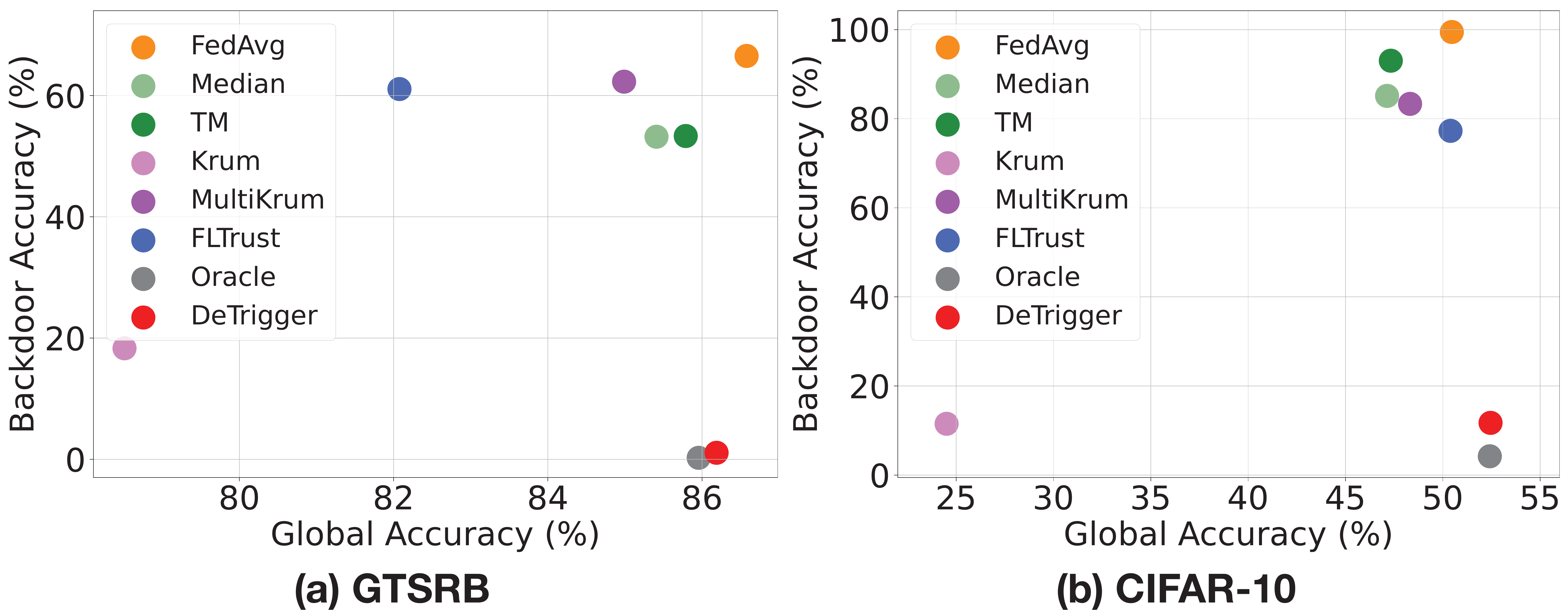}
    \vspace{-2ex}
    \caption{Overall global model accuracy v.s. backdoor attack accuracy across different baselines.}
    \label{fig:overall}
\end{figure}

We begin our evaluations by presenting the overall performance of \system{} to understand how well the global models in \system{} perform while mitigating backdoor attacks. Figure~\ref{fig:overall} plots the global model accuracy and backdoor accuracy for \system{} and different comparison baselines with two different datasets. The backdoor accuracy denotes the attack success rate when a trigger is present in the input, while global accuracy shows the classification performance on unaltered samples. In this context, an ideal approach would appear in the bottom-right corner of the plot, signifying low backdoor accuracy (effective attack mitigation) and high global accuracy (preserved model performance)

As the results show, \system{} consistently shows superior performance in achieving a balance between these two performance metrics. For the GTSRB dataset (Figure~\ref{fig:overall} (a)), while FedAvg demonstrates high global model accuracy, it is significantly susceptible to backdoor attacks. Similarly, methods such as FLTrust and Median slightly reduce backdoor accuracy, but these come at the cost of degraded global model performance. MultiKrum and Trimmed Mean (TM) show improvements in robustness, but their overall global model fails to reach the level of \system{}.

Similarly, with the CIFAR-10 dataset (Figure~\ref{fig:overall} (b)), \system{} outperforms all baselines. The baselines achieve marginal improvements in backdoor resistance but exhibit degradation in global model accuracy. In contrast, \system{} significantly suppresses backdoor accuracy while maintaining competitive global model performance, narrowing the gap toward the ideal results observed with Oracle-level defenses. 

One interesting observation we make is that \system{} achieves higher global model accuracy compared to the Oracle baseline for both datasets. This is due to the fact that, unlike Oracle, which completely excludes attacker models from aggregation, \system{} carefully prunes only the backdoor-related information while retaining and leveraging the benign knowledge of a compromised model. This showcases the effectiveness of our approach in preserving valuable model updates while mitigating backdoor attacks.

\subsection{Computational Efficiency of \system{}}

\begin{figure}
    \centering 
    \includegraphics[width=0.95\linewidth]{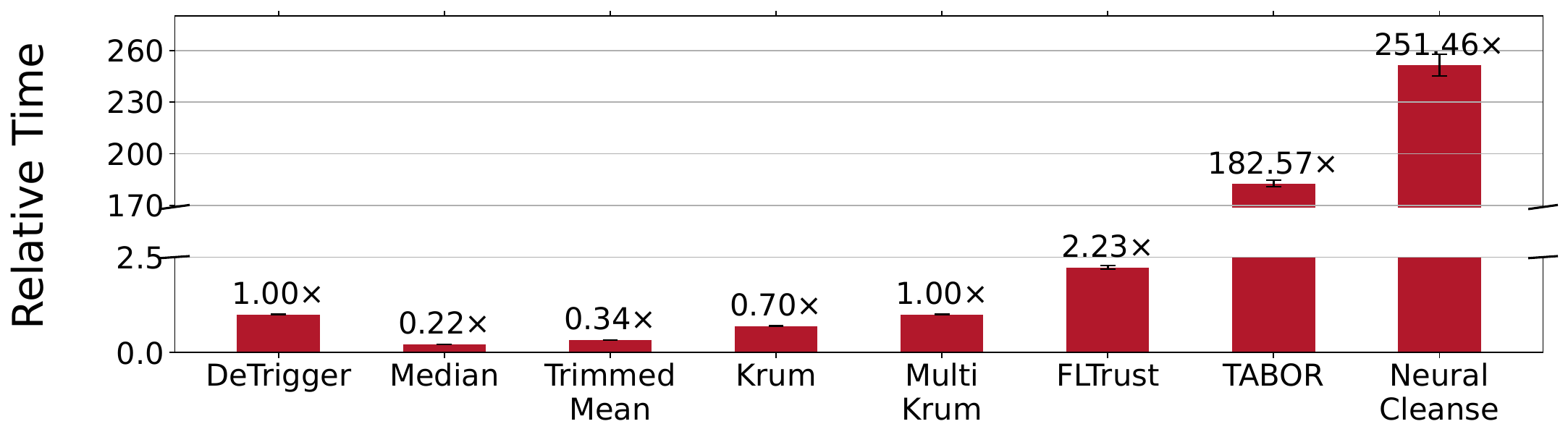}
    \vspace{-2ex}
    \caption{Elapsed time per federated training round with respect to different defense schemes normalized to the performance of \system{}.}
    \label{fig:time-complexity}
\end{figure}

As highlighted earlier, the time required to validate models in mitigating backdoor attacks is a critical factor as it can introduce delays in the overall federated training process. Figure~\ref{fig:time-complexity} presents the elapsed time for a single federated learning round with 50 clients, including 25 backdoor attackers, averaged over 100 trials. Here, the performance of all comparison methods is normalized to the latency of \system{}, which is $\sim$0.71~sec. The plots show that defense schemes based on safe aggregation with statistical priors, such as Median and Trimmed Mean (TM), exhibited relatively low time complexity, being 4.56$\times$ and 2.94$\times$ faster than \system{}, respectively. On the other hand, methods such as Krum, MultiKrum, and FLTrust, which compute the similarity between model updates, showed slightly higher computational costs due to the additional similarity calculations required in their schemes. While their latency is at a practically acceptable level (some even faster than \system{}) we show in the following evaluations that this comes at the price of failing to mitigate the backdoor attack in many cases properly.

Furthermore, Figure~\ref{fig:time-complexity} indicates that the optimization and training processes in TABOR and Neural Cleanse, which aim to detect backdoor models and extract triggers (similar to \system{}), show significantly higher computational latency, being 182.57$\times$, and 251.46$\times$ higher than \system{}, respectively. Given that the system will encounter such latency at every federated learning round, and also that this latency will increase with an increasing number of clients, we see the computational latency being a significant issue for practically adopting these previously proposed schemes. 

\begin{figure}[!t]
    \centering
    \includegraphics[width=0.9\linewidth]{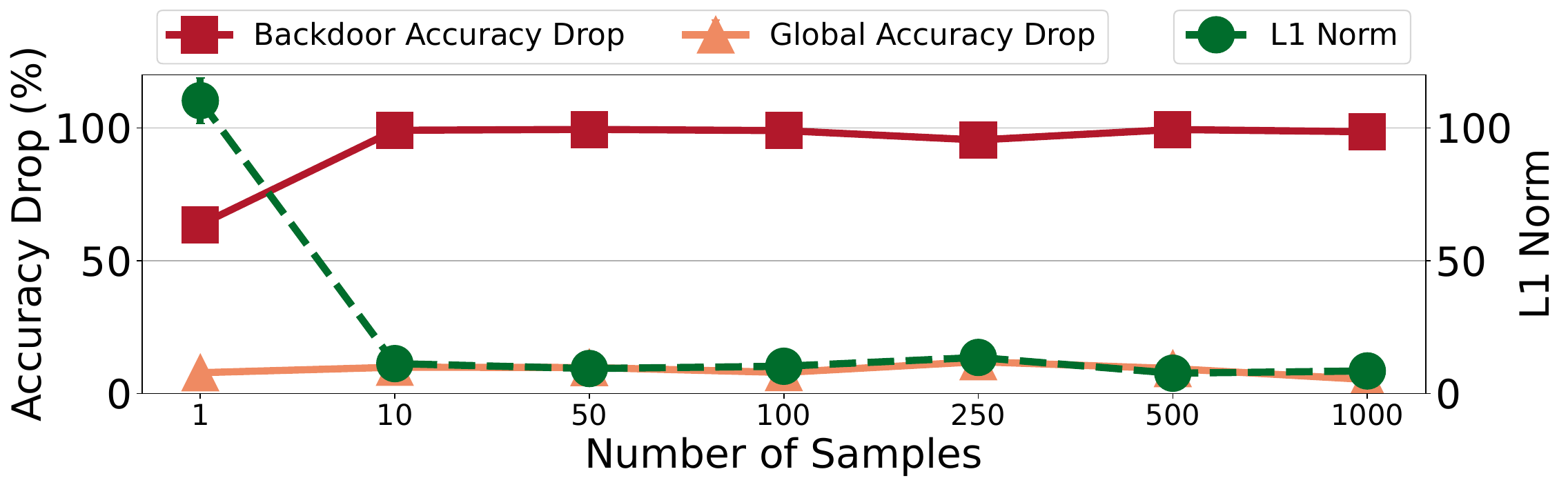}
    \vspace{-2ex}
    \caption{L1-norm of ground truth and inferred trigger along with the accuracy drop for the backdoor and global model with varying the number of validation samples at the server.}
    \label{fig:varying-nsample-l1norm}
\end{figure}

Recall that \system{} computes input layer gradients using unaltered validation samples at the server; thus, its performance relies on the availability of these samples. We perform an evaluation to examine \system{}'s performance with varying validation sample quantities for gradient computation.

Figure~\ref{fig:varying-nsample-l1norm} shows the global model and backdoor attack accuracy drop rates, and the L1 norm between the ground truth and \system{}-predicted trigger patterns for validation sample sizes ranging from 1 to 1000. When the size of the validation set is extremely low (e.g., 1), the L1 norm exhibited higher error, and the backdoor defense performance degraded ($\sim$60\% block rate). However, with just 10 validation samples (randomly selected from a set of 1,000), \system{} achieved significantly improved trigger prediction quality and a noticeable increase in the attack block rate.

\subsection{Impact of Backdoor Knowledge Pruning}
\begin{figure}[!t]
    \centering
    \includegraphics[width=0.95\linewidth]{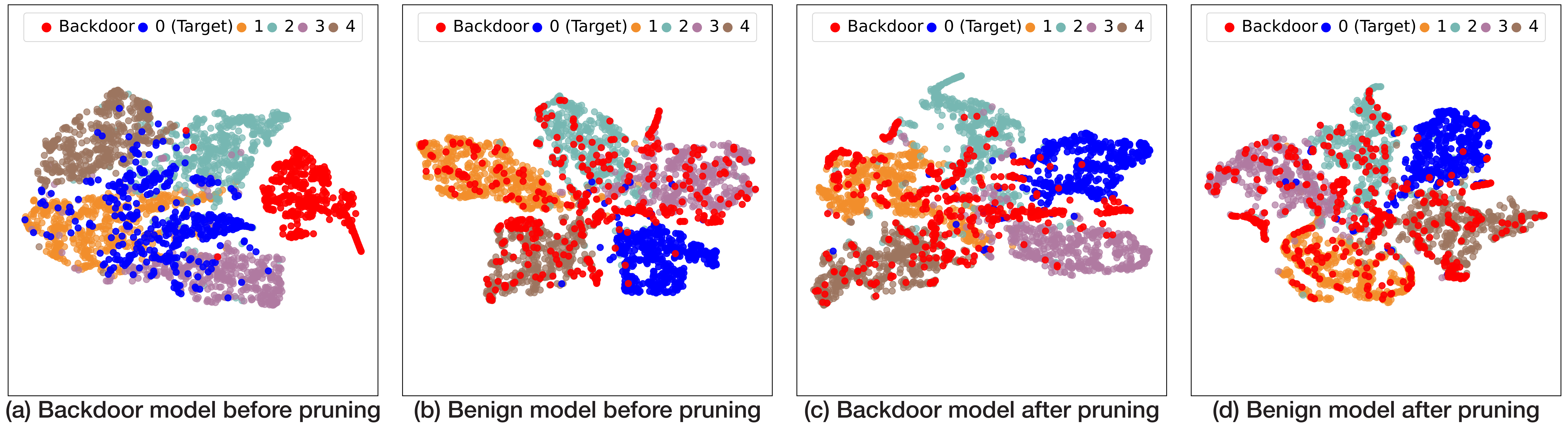}
    \caption{Visualization of representation space before and after weight pruning is applied on backdoor and benign models.}
    \label{fig:pruning-vis}
\end{figure}    

\begin{figure}[!t]
    \centering
    \includegraphics[width=\linewidth]{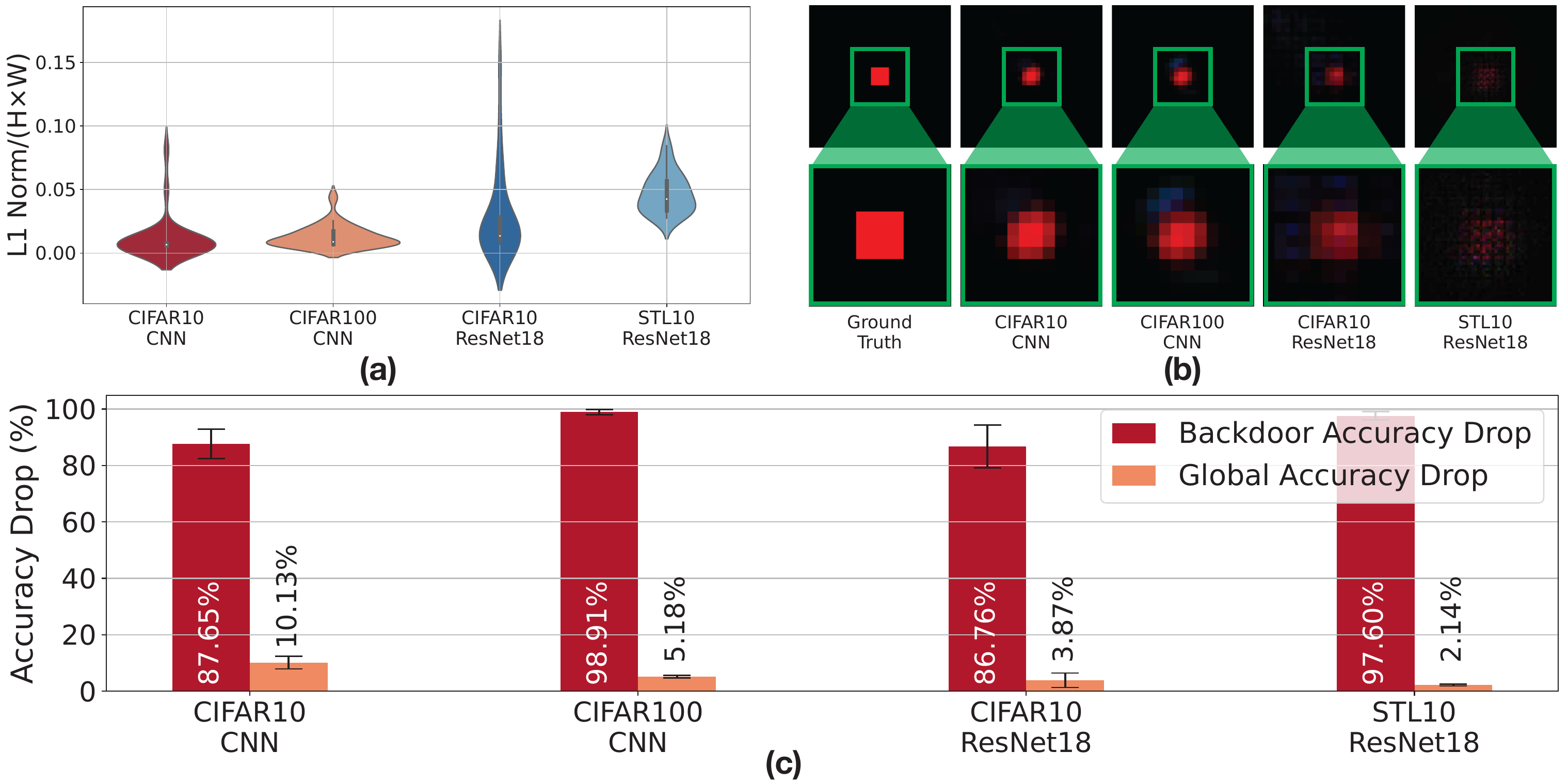}
    \vspace{-3ex}
    \caption{(a) L1-norm with respect to input size for different dataset/model configurations. (b) Sample triggers extracted for different dataset/model configurations. (c) Accuracy drop of backdoor and global model accuracy for different dataset/model configurations.}
    \label{fig:varyingcase}
\end{figure}



To evaluate the effectiveness of backdoor knowledge pruning, we visualized the learned representations of both benign and backdoor models (GTSRB dataset and CNN configuration) using t-SNE~\cite{van2008visualizing}. As t-SNE maps data to arbitrary spaces; thus, absolute locations or distances are not directly comparable across figures. For clarity, we only include the top five labels, including the attack target label (from 43 classes).

Figures~\ref{fig:pruning-vis} (a) and (b) illustrate the learned representations \textit{prior to} pruning for backdoor and benign models, respectively. As shown, the backdoor samples (red dots) in Figure~\ref{fig:pruning-vis} (a) are clustered separately from the normal samples due to the attack constructing an independent decision boundary (for intentional misclassification), whereas in Figure~\ref{fig:pruning-vis} (b), the benign model maps the backdoor samples with their original labels rather than the target attack label. This indicates that a benign model is not affected by the attack trigger.

We present the representations of backdoor-affected and benign models \textit{after} the pruning process using the inferred trigger in Figures~\ref{fig:pruning-vis} (c) and (d), respectively. Figure~\ref{fig:pruning-vis} (c) shows that pruning disrupts the red cluster associated with the backdoor, redistributing these samples toward the normal data feature space. Note that in some cases, \system{} may occasionally misclassify a benign model as compromised and apply pruning. However, Figure~\ref{fig:pruning-vis} (d) demonstrates that such unintended pruning has minimal impact on benign model behavior, as unrelated knowledge (e.g., labels 0-4) remains intact post-pruning. 

\subsection{Performance Across Varying Datasets and Model Characteristics}
Next, we evaluate \system{}'s performance across diverse dataset-model configurations, focusing on (i) label types, (ii) input resolutions, and (iii) model characteristics. 

Figure~\ref{fig:varyingcase} (a) presents the L1 norm between the ground truth trigger pattern and the inferred trigger across various configurations with 75 benign clients and 25 backdoor attackers total of 100 participants. To ensure fair comparisons across resolutions, the L1 norm is normalized by dividing it by the spatial dimension (\(H \times W\)), as input data dimensions affect the norm. The results here show stable inferred trigger quality across configurations, with a minor exception in the STL10-ResNet18 setting, which combines high-resolution input with a relatively complex model.

Next, Figure~\ref{fig:varyingcase} (b) visualizes the reverse-engineered triggers for different dataset-model configurations. Note that for STL10-ResNet18, a clear trigger pattern is less evident, suggesting possible performance degradation in scenarios involving high-resolution data and complex models. Nevertheless, despite imperfect trigger extraction from a human perception perspective, \system{} effectively removes trigger-related information during its pruning phase while retaining benign model knowledge. Specifically, Figure~\ref{fig:varyingcase} (c) shows the accuracy drop rate for the backdoor attack and global model before and after pruning. We can notice that the global model accuracy remains largely unaffected, while backdoor accuracy is reduced by up to 98.90\% in the CIFAR100-CNN case. This aligns with findings by Wang et al.~\cite{wang2019neural}, who noted that backdoors can still exploit imperfect patterns. Overall, these results demonstrate that \system{} is flexible enough to support a variety of dataset and model configurations within federated learning operations.

\subsection{Scalability of \system{}}

\begin{figure}[!t]
    \centering
    \includegraphics[width=.675\linewidth]{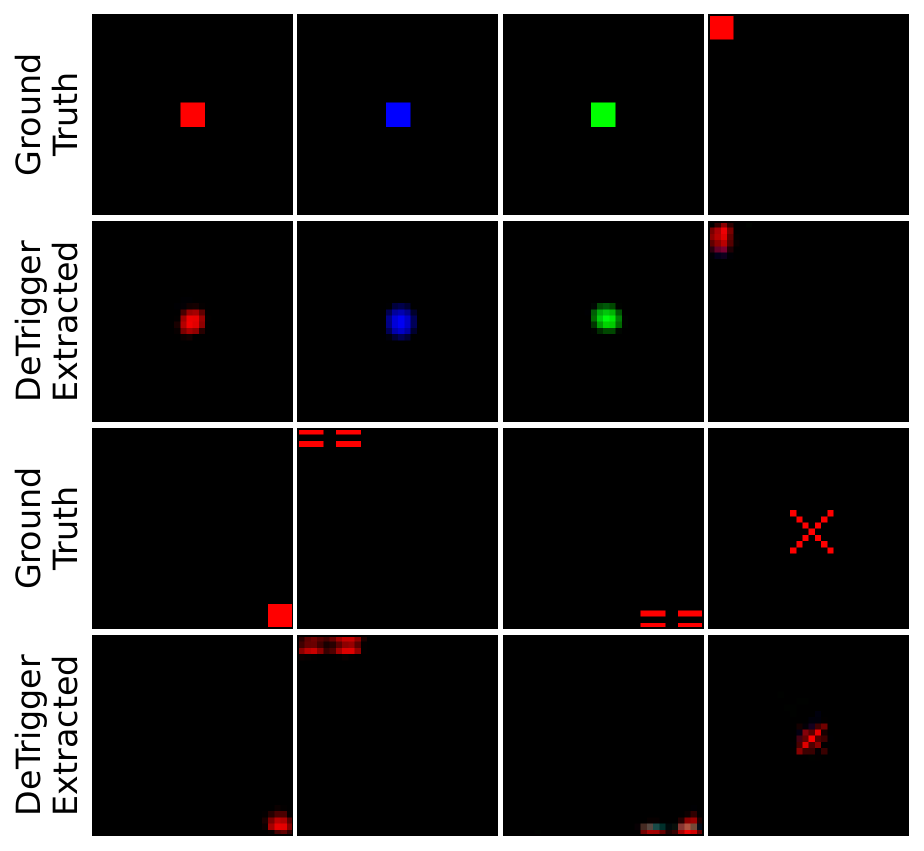}
    \caption{Extracted sample triggers with varying color, location, and shape of the backdoor trigger.}
    \label{fig:varying-trigger}
\end{figure}

Finally, we evaluate \system{}'s scalability on different backdoor trigger patterns and the increasing number of participating clients in the federated learning network.

\subsubsection{Different Backdoor Trigger Patterns}
Trigger patterns for backdoor attacks can vary widely. While small patterns appended to the original input are common characteristics, their visual characteristics, such as shape and color, can differ significantly. To evaluate the scalability and robustness of \system{} against diverse backdoor triggers, we conducted experiments using eight different trigger types, visualized at the top of Figure~\ref{fig:varying-trigger} leveraging the 2-layered CIFAR10 dataset and CNN model. As shown in the bottom of the figure, \system{} successfully extracts the triggers across various shapes and colors, albeit not perfectly.

The L1 norm between the ground truth and inferred trigger patterns, shown in Figure~\ref{fig:varying-trigger-l1norm}, indicates that non-continuous pixel patterns (e.g., the final three triggers in Figure~\ref{fig:varying-trigger}) tend to show increased quantitative detection error. However, at a system level, Figure~\ref{fig:varying-trigger-asracc} demonstrates that even imperfect trigger extractions, when integrated into the full \system{} pipeline, effectively maintain a highly accurate global model while significantly suppressing backdoor attacks.

\begin{figure}[t]
    \centering
    \includegraphics[width=0.9\linewidth]{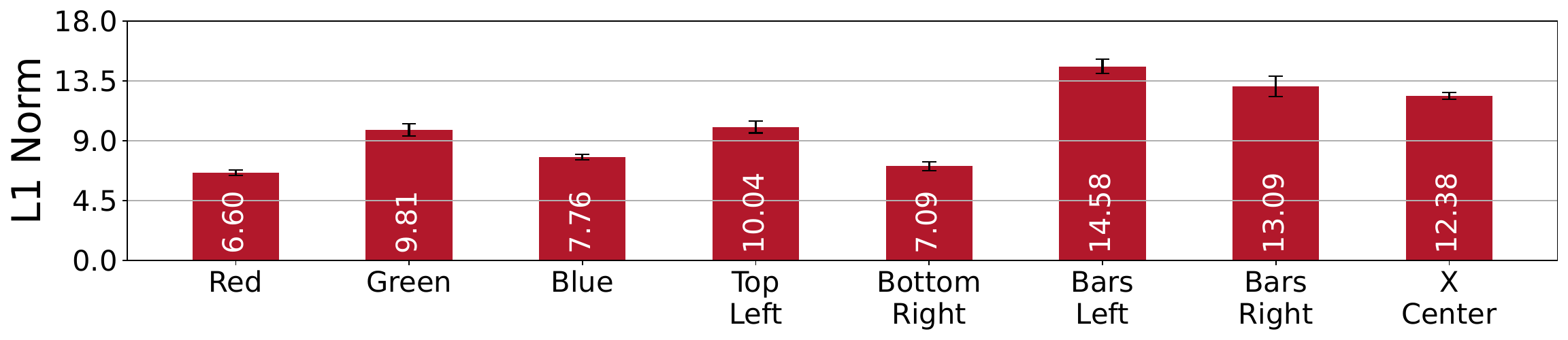}
    \vspace{-2ex}
    \caption{L1-norm with varying color, location, and shape of the backdoor trigger.}
    \label{fig:varying-trigger-l1norm}
\end{figure}

\begin{figure}[t]
    \centering
    \includegraphics[width=0.9\linewidth]{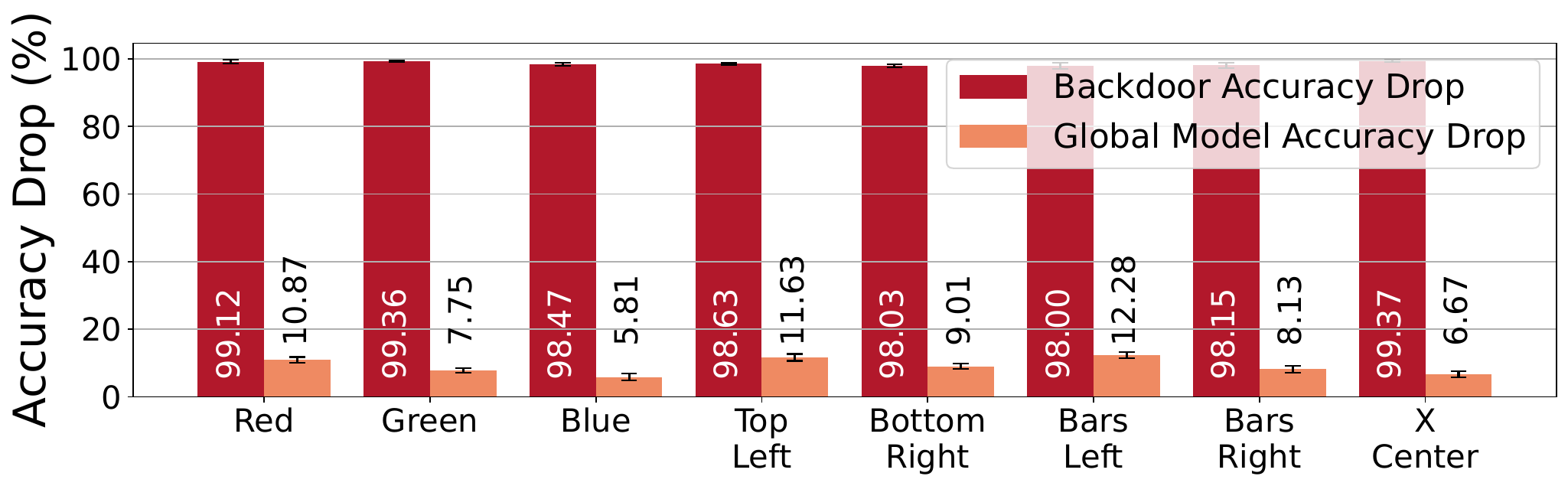}
    \vspace{-2ex}
    \caption{Accuracy drop for the backdoor task and main task for different colors, locations, and shapes of the backdoor trigger.}
    \label{fig:varying-trigger-asracc}
\end{figure}

\subsubsection{Number of Participating Clients}

We now examine how \system{} scales with the increasing number of clients in its federated learning network. Here, we focus on the computational overhead of dealing with the increased number of updated models that are collected at the server. This is particularly important given that long latencies lead to increased intervals between federated learning rounds, which in turn translates to prolonged model convergence. 

Our results with the CIFAR10 dataset and 2-layered CNN model plotted in Figure~\ref{fig:varying-nclient-time} suggest that, as expected, the overall computation time for \system{}'s operations increases with increasing clients. However, the latency for \system{} is significantly lower compared to previously proposed alternatives such as TABOR and Neural Cleanse. This results suggests that \system{} is an efficient and scalable solution for addressing backdoor attacks in FL systems.

\begin{figure}[t]
    \centering
    \includegraphics[width=0.9\linewidth]{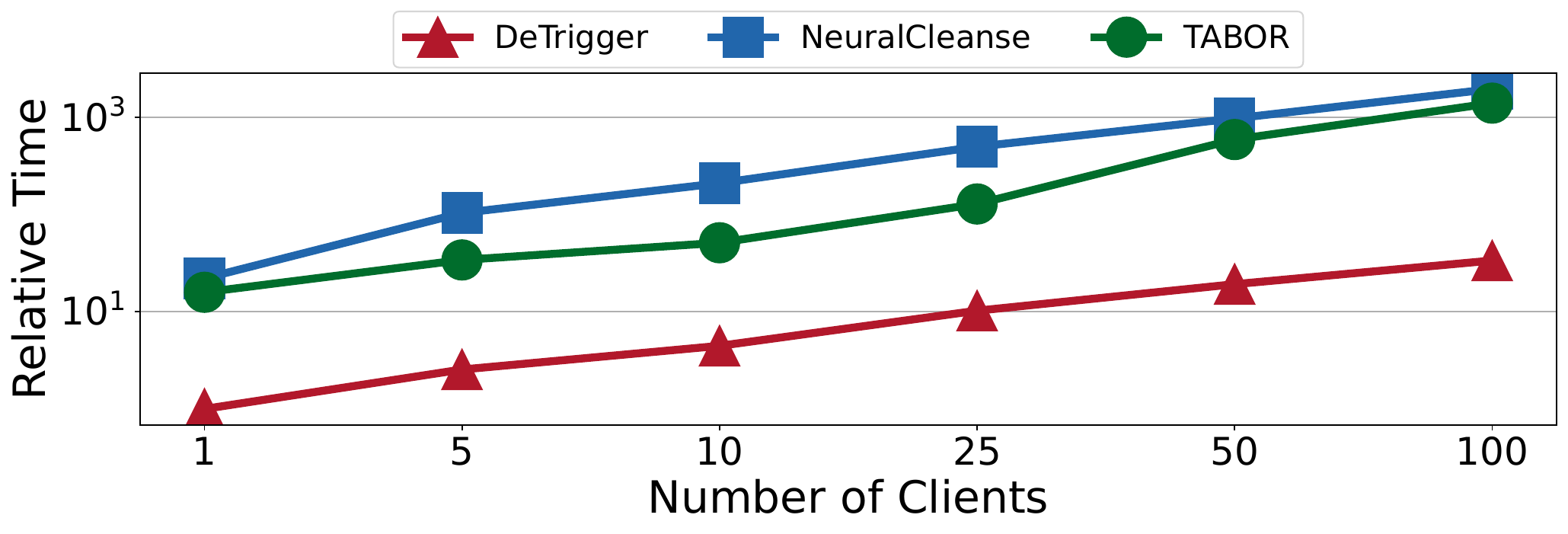}
    \vspace{-2ex}
    \caption{Normalized latency for different backdoor detection schemes with varying number of clients.}
    \label{fig:varying-nclient-time}
\end{figure}
\section{Discussion}
\label{sec:discussion}

Based on our experiences in designing and evaluating \system{}, we discuss the limitations of our current research and suggest directions for future work.

\noindent\textbf{$\bullet$ Understanding backdoor attack via gradients.} 
In this paper, we explored the relationship between backdoor and adversarial attacks using gradient analysis. We also introduced the temperature scaling trick, offering a novel perspective on the decision boundary of backdoor models. These analyses provide valuable insights that enhance the understanding of backdoor attacks. We hope the insights presented here will serve as a foundation for further research and inspire new discussions on backdoor vulnerabilities in neural networks.

\noindent\textbf{$\bullet$ Exploiting advanced adversarial attacks.} 
Our work primarily seeks to emphasize and validate the feasibility of leveraging adversarial attack concepts to mitigate backdoor attacks in federated learning. Accordingly, \system{} utilizes a straightforward gradient-based adversarial attack. We acknowledge that prior studies have proposed more advanced adversarial attack techniques. We believe that incorporating these approaches could further enhance the effectiveness of our defense mechanism.

\noindent\textbf{$\bullet$ Adaptive attack against \system{}.} 
To further strengthen the federated learning framework, it is essential to explore its limitations and conduct stress testing of the defense mechanism. While our work demonstrates the effectiveness of leveraging adversarial attacks to mitigate backdoor attacks, we also consider potential adaptive attacks that could be designed to evade \system{}. For example, an adaptive attack might employ triggers with high total variation to bypass detection, as \system{} uses total variation metrics to identify backdoor updates. Importantly, the global model accuracy did not significantly degrade from the pruning process, indicating that \system{} can adapt to counter these attacks by expanding its detection criteria to account for a broader range of potential triggers. Furthermore, adjusting the total variation $TV$ threshold would allow \system{} to maintain its effectiveness against adaptive threats.

\noindent\textbf{$\bullet$ Attack pattern limitations.} 
Although \system{} has proven effective in mitigating backdoor attacks that embed spatially concentrated triggers in input data, its generalizability to dispersed triggers or non-trigger-based backdoor strategies remains an open question. Exploring this aspect presents an intriguing direction for future research.
Nevertheless, previous works in backdoor attacks suggest that a trigger pattern concentrated on a single position is common and widely used. A major benefit of exploiting such patterns is that the trigger is not prominent in the input data. Thus, we see \system{} being a useful tool for such attacks, forcing a more easily noticeable attack trigger when issuing a backdoor attack.

\section{Conclusion}
\label{sec:conclusion}

In this paper, we introduced \system{}, a backdoor-robust federated learning framework designed to detect and mitigate backdoor attacks by leveraging adversarial attack methodologies. Through both theoretical and empirical gradient analysis and temperature scaling, \system{} effectively isolates trigger patterns, enabling model weight pruning for the removal of backdoor activations while retaining benign knowledge within the global model. Our extensive evaluations demonstrate that \system{} not only achieves significant speed improvements over traditional backdoor defenses but also preserves model accuracy and mitigates attack effectiveness by up to 98.9\%. Additionally, through extensive evaluations using four widely-used public datasets, we explored the scalability of \system{} across diverse settings, confirming its adaptability to varying model complexities, label sizes, and data resolutions. By combining efficiency with precision, \system{} sets a foundation for secure and scalable federated learning. 

\bibliographystyle{ieeetr}
\bibliography{reference, eis-lab}
\appendix

\subsection{Proof for Lemma~\ref{lem:initial_derivative}}
The proof of Lemma~\ref{lem:initial_derivative} is as follow:
\begin{proof}
Define $f_x(\alpha):=\langle \nabla_x z_{y_t}(p(\alpha)), \Delta x\rangle$.
Then $f_x'(\alpha)=\Delta x^\top H_{z_{y_t}}(p(\alpha))\Delta x \le \kappa\|\Delta x\|^2$ by Assumption~\ref{ass:dominance_curvature}.
Hence $f_x(\alpha)\le f_x(0)+\alpha\kappa\|\Delta x\|^2$. Integrating over $\alpha\in[0,1]$ gives
\[
\Delta_{BD}(x)=\int_0^1 f_x(\alpha)\,d\alpha
\le f_x(0)+\frac{\kappa}{2}\|\Delta x\|^2,
\]
which implies $f_x(0)\ge \Delta_{BD}(x)-\frac{\kappa}{2}\|\Delta x\|^2$.
Taking expectations over $x$ yields the claim.
\end{proof}

\subsection{Proof for Theorem~\ref{thm:expected_alignment}}
\begin{proof}
The cross-entropy gradient expansion yields
\[
g(x)=(1-p_{y_t}(x))\nabla_x z_{y_t}(x)-\sum_{j\neq y_t} p_j(x)\nabla_x z_j(x).
\]
Taking inner product with $\Delta x$ and expectation over $x$,
\begin{align*}
\mathbb{E}[\langle g,\Delta x\rangle]
&=
\mathbb{E}\big[(1-p_{y_t})\langle \nabla z_{y_t},\Delta x\rangle\big]
-\mathbb{E}\Big[\sum_{j\neq y_t} p_j \langle \nabla z_j,\Delta x\rangle\Big].
\end{align*}
Since $p_{y_t}(x)\le \rho$, we have $(1-p_{y_t}(x))\ge (1-\rho)$ and thus
\[
\mathbb{E}\big[(1-p_{y_t})\langle \nabla z_{y_t},\Delta x\rangle\big]
\ge (1-\rho)\,\mathbb{E}[\langle \nabla z_{y_t},\Delta x\rangle]
\ge (1-\rho)\mu_{\text{target}},
\]
where the last inequality follows from Lemma~\ref{lem:initial_derivative}.
For the non-target term,
\[
-\mathbb{E}\Big[\sum_{j\neq y_t} p_j \langle \nabla z_j,\Delta x\rangle\Big]
\ge
-\mathbb{E}\Big[\sum_{j\neq y_t} p_j \big|\langle \nabla z_j,\Delta x\rangle\big|\Big]
\ge -\beta
\]
by Assumption~\ref{ass:interference}. Combining the bounds proves the theorem.
\end{proof}

\subsection{Proof for Theorem~\ref{thm:cosine_convergence}}
\begin{proof}
By Assumption~\ref{ass:signal_noise}, $\hat g=\hat c\,T_M+\bar v$, where $\hat c=\frac{1}{N}\sum_i c_{x_i}$ and $\bar v=\frac{1}{N}\sum_i v_{x_i}$.
Since each $v_{x_i}\perp T_M$, we have $\bar v\perp T_M$ and thus
\[
\quad \quad \cos(\hat g,T_M)
=
\frac{\langle \hat g,T_M\rangle}{\|\hat g\|\|T_M\|}
\]
\[=
\frac{\hat c\|T_M\|}{\sqrt{\hat c^2\|T_M\|^2+\|\bar v\|^2}}
\]
\[
=
\frac{1}{\sqrt{1+u}},
\]
\[
\quad
u:=\frac{\|\bar v\|^2}{\hat c^2\|T_M\|^2}.
\]
Using the inequality $(1+u)^{-1/2}\ge 1-\frac{u}{2}$ for $u\ge 0$,
\[
\cos(\hat g,T_M)\ge 1-\frac{\|\bar v\|^2}{2\hat c^2\|T_M\|^2}.
\]
Standard concentration for sub-Gaussian vectors implies that, with probability at least $1-\delta$,
\[
\|\bar v\|^2 \le C_2 \frac{\sigma^2\,(d+\log(1/\delta))}{N}.
\]
On the same event, if $\hat c \ge \bar c/2$, substituting the bounds yields the stated inequality (absorbing constants into $C_1$).
\end{proof}



\end{document}